\documentclass{nle}

\usepackage[pdftex]{graphicx}
\graphicspath{{../pdf/}{../jpeg/}}
\DeclareGraphicsExtensions{.pdf,.jpeg,.png}
\pdfoutput=1
\usepackage{subfigure}
\usepackage{amsmath}
\usepackage{amssymb}
\usepackage{graphicx}
\usepackage{multirow}
\usepackage[multidot]{grffile}
\usepackage{graphicx}
\usepackage{subfigure}
\usepackage{multirow}
\usepackage{multicol}
\usepackage{color}
\usepackage{comment}
\usepackage{verbatim}
\usepackage{url}
\usepackage[normalem]{ulem}
\usepackage[ruled,linesnumbered]{algorithm2e}
\usepackage{stfloats}
\usepackage{enumerate}
\usepackage{enumitem}
\usepackage{underscore}
\usepackage{ulem}
\usepackage{verbatim}
\usepackage{mathtools}
\usepackage{multirow}
\usepackage{booktabs}

\title[Lifelong Topic Model for Document Summarization]
      {Joint Lifelong Topic Model and Manifold Ranking for Document Summarization}
\author[J. Lin {\it et al.}\ ]
       {J\ls I\ls A\ls N\ls Y\ls I\ls N\ls G\quad L\ls I\ls N\textsuperscript{1,2}, R\ls U\ls I\quad L\ls I\ls U\textsuperscript{1,2,3}, Q\ls U\ls A\ls N\ls Y\ls E\quad J\ls I\ls A\textsuperscript{1,2}\\
        \textsuperscript{1}State Key Laboratory of Software Development Environment, Beihang University, Beijing, China\\
        \textsuperscript{2}School of Computer Science and Engineering, Beihang University, Beijing, China\\
        \textsuperscript{3}corresponding author: lr@buaa.edu.cn}

\pagerange{\pageref{firstpage}--\pageref{lastpage}}
\pubyear{1998}

\begin{document}

\label{firstpage}
\maketitle

\begin{abstract}
Due to the manifold ranking method has a significant effect on the ranking of unknown data based on known data by using a weighted network, many researchers use the manifold ranking method to solve the document summarization task. However, their models only consider the original features but ignore the semantic features of sentences when they construct the weighted networks for the manifold ranking method. To solve this problem, we proposed two improved models based on the manifold ranking method. One is combining the topic model and manifold ranking method ({\em JTMMR}) to solve the document summarization task. This model not only uses the original feature, but also uses the semantic feature to represent the document, which can improve the accuracy of the manifold ranking method. The other one is combining the lifelong topic model and manifold ranking method ({\em JLTMMR}). On the basis of the {\em JTMMR}, this model adds the constraint of knowledge to improve the quality of the topic. At the same time, we also add the constraint of the relationship between documents to dig out a better document semantic features. The {\em JTMMR} model can improve the effect of the manifold ranking method by using the better semantic feature. Experiments show that our models can achieve a better result than other baseline models for multi-document summarization task. At the same time, our models also have a good performance on the single document summarization task. After combining with a few basic surface features, our model significantly outperforms some model based on deep learning in recent years. After that, we also do an exploring work for lifelong machine learning by analyzing the effect of adding feedback. Experiments show that the effect of adding feedback to our model is significant.
\end{abstract}
\section{Introduction}
With the rapid development of the Internet, the Internet has become an important carrier of current information. The emergence of social networks allows people to publish and share information on the Internet, such as current news, blog posts, personal feeling, etc. According to the statistics of IDC, the amount of Internet data has jumped to the ZB level ($1ZB=2^{40}GB$) since 2010, and it is expected to reach 40ZB by 2020. At the same time, people are increasingly relying on search engines to obtain the required Internet information, but search engines cannot effectively solve the problem of information overload. On the Internet, there are a large number of web pages with the same theme or similar content. Especially in news, many media often have similar attitudes toward the same thing, which leads to the content from different media is roughly the same. When a user enters a keyword on a search engine, the search engine often returns redundant, one-sided and contaminated information. Therefore, it is important to figure out how to extract common key information and information that users are interested in from the same type of document under the same topic, and retain their own difference information, which can help users to get the useful information on the Internet more effectively.

Document summarization technology is an important application field of natural language processing. Its main purpose is to generate a summary by automatically write and rewrite original documents. This technology can help people to comprehend the content of natural language documents in a general way and can present important information in massive documents more quickly and accurately, which can solve the problem of information overload. With the development of document summarization, it has also been widely applied to various fields, including text classification \cite{shen2004web}, information retrieval \cite{glavavs2014event}, question answering system \cite{morita2012query}, etc. These applications can make search engines faster and better present the results that users want.

With the increasing amount of data, document summarization models based on a single batch of corpus training have been unable to adapt to the needs of users because the time and space complexity of models are often more than $O(n^{2})$. For massive data, such complexity often means inefficient, and in some application scenarios it is unacceptable. These years, the method of document summarization has also been improved. For example, researchers proposed many methods based on statistical features \cite{ferreira2014multi}, lexical chains \cite{li2007query} and graph ranking \cite{litvak2008graph} to generate summary. With the rise of deep learning, there are also many document summarization methods by using deep learning \cite{zhang2017multiview,li2017cascaded,li2017salience}, which can make the effect of document summarization increasing. However, these methods are based on the training of a single batch of corpus and do not adapt well to the rapid development of Internet data.

In order to solve the above problem, a method based on lifelong machine learning is used here, so that the machine can use a "brain" like a human, continuously accumulate knowledge, and apply knowledge to the following tasks to improve the effect of subsequent tasks. The method of lifelong machine learning is more in line with the current explosive growth in the volume of information data on the Internet. It is consistent with human learning because the method can continue to run continuously with an increasing amount of information, and generate summary more accurately, efficiently, and comprehensively. Therefore, lifelong machine learning methods are necessary for document summarization tasks.

We first combined the non-negative matrix factorization topic model and the manifold ranking model (JTMMR) to modify the generation of weighted matrix by introducing deep semantic features into the manifold ranking, so as to improve the accuracy of sentence ranking scores by manifold ranking. Then, we propose a lifelong topic model to replace the non-negative matrix factorization topic model in the JTMMR model. It can not only constrain the generation of the topic-to-word relationship to improve the quality of topic but also join the constraints of document relationships to mine better document semantic features by introducing knowledge to the JTMMR model. Combined with the lifelong topic model and the manifold ranking model (JLTMMR), using better deep semantic features, construct a weighted matrix of manifold sorting to improve the effect of manifold sorting on the significance of sentence scoring. Extensive experiments prove that our models can combine the advantage of topic model and manifold ranking method to achieve a better result.

In a nutshell, the contributions are threefold as below:

(1) JTMMR

In the study of the algorithm, we present our proposed joint topic model and manifold ranking method (JTMMR) for document summarization. It combines the advantage of semantic features and original features of documents. It has been proved to enhance the quality of summary with 1\% to 2\% on ROUGE F-measure.

(2) Lifelong Topic Model

We propose a lifelong topic model by adding the constraint of knowledge and document relationship. It can improve the performance of topic model in the short text environment and make the topic model more suitable for the big data.

(3) JLTMMR

We use the lifelong topic model to replace the non-negative matrix factorization topic model in the JTMMR model, and then combine the lifelong topic model and manifold ranking methods (JLTMMR) to generate the summary. In addition, we add the statistical features (SF) to the JLTMMR model in the calculation of the sentence saliency. At analysis of measures, our method is better than most of the baseline model in multi-document summarization task and achieve the third place results on ROUGE-2 in single-document summarization task.

The rest of this paper is organized as follows. Section 2 gives an introduction about related work. The model of the JTMMR is described in Section 3. In Section 4, we propose the JLTMMR model by changing topic model in JTMMR model to lifelong topic model. Section 5 provides a numerical example and compares the quality of summary among all alternatives. The conclusions are provided in Section 6.
\\

\section{Related work}
\subsection{Non-negative Matrix Factorization Based Topic Model}
The non-negative matrix factorization (NMF) method \cite{lee1999learning} assumes that the original matrix can be decomposed into the product of two nonnegative matrices. It always uses to mine the semantic topic features. Suppose there is a text dataset containing $N$ documents. Then the document can be obtained by word segmentation, removal of stopwords, and tense changing. After that, we can get a vocabulary containing $M$ words. Thus, each document consists of an M-dimensional vector, and the weight of each vector is calculated by TFIDF \cite{salton1988term}. The document dataset can be represented as a word-to-document matrix $D$ with a dimension of $M*N$. In the decomposition process, we hope to split $D$ into the product of the two matrices $U$ and $V$, while limiting the values of all elements in $U$ and $V$ to be non-negative. The dimension of the $U$ matrix is $M*K$, which represents the distribution of words under the topic. The dimension of the $V$ matrix is $K*N$, which represents the distribution of topics under the document. After the non-negative matrix factorization, the document matrix is represented by the implicit semantic features. This will not only reduce the amount of computation through dimensionality reduction, but also excavate the implicit semantic features of the document.
\subsection{Lifelong Machine Learning}
The lifelong machine learning method \cite{silver2013lifelong} is an advanced machine learning method that can learn continuously, accumulate the knowledge learned in the previous task and use the knowledge to help the learning of future tasks. Given a set of tasks $T={T_1,T_1,...,T_N}$, corresponding to the datasets $D=D_1,D_2,...,D_N$ for each task. Between each task, there is a shared knowledge base. Then for the next task $T_{N+1}$, it will use the knowledge learned by the previous $N$ tasks and stored on the shared knowledge base to training the model of task $T_{N+1}$. After the training is completed, the knowledge learned this time is stored and updated in the knowledge base. The process of updating knowledge can be continuously checked, questioned, and tapped to explore higher levels of knowledge. In this process, models (learners) have more and more knowledge and can learn efficiently.

Learning ability is one of the hallmarks of human intelligence. However, there are many machine learning methods do not try to preserve the learned knowledge and use it in future learning. Although this kind of isolated machine learning method has been very successful, it requires a large number of training examples and it is only suitable for clearly defined and narrow tasks \cite{chen2016lifelong}. In contrast, as humans, we can learn new knowledge by using little data because we have accumulated a lot of knowledge in the past. Lifelong machine learning is designed to enable the machine to have the same learning ability as human beings to enhance the learning effect.
It is clearly that lifelong machine learning has the following core elements: (1) It can be carried out continuously; (2) It can excavate knowledge and store knowledge; (3) It can use the knowledge learned in the past to help the later learning. The lifelong machine learning method has now been applied to supervised learning tasks \cite{chen2015lifelong,ruvolo2013ella}, unsupervised learning tasks \cite{chen2014topic}, semi-supervised learning tasks \cite{carlson2010toward,movshovitz2012bootstrapping} and reinforcement learning tasks \cite{fernandez2013learning}.
\subsection{Manifold Ranking}
The manifold ranking method \cite{zhou2004learning,zhou2004ranking} is a general sorting algorithm. The algorithm was originally used to sort data points to determine its underlying manifold structure. Manifolding is based on the following two assumptions: (1) Adjacent points have approximate ranking scores; (2) Points on the same type of structure (e.g., belonging to the same cluster or the same manifold structure) also have approximate ranking scores. The detail of manifold ranking is to form a weighted network on the dataset by assigning a positive score to each known related point and assigning a zero score to the data points of the remaining unknown scores. Then all points pass their scores through the weighted network to the neighboring nodes. By repeating this propagation process until the global stable state is reached, the final ranking score for all points is obtained. It is mainly applied to text categorization \cite{zhou2004learning}, document sorting \cite{zhou2004ranking}, query recommendation \cite{zhu2011query}, handwritten digit recognition \cite{zhou2004learning} and image saliency detection \cite{yang2013saliency}.
\subsection{Document Summarization}
For document summarization technology, there are different divisions according to their different characteristics. Normally, there are two divisions most commonly used by researches. One is dividing document summarization into single document summarization and multi-document summarization according to the number of input documents; the other is dividing into extractive document summarization and abstractive document summarization based on the relationship between generated summaries and original texts. The main purpose of extractive document summarization is to generate the final summary by directly extracting the sentence in the document. The main reason is that the sentences in the document are consistent, coherent, smooth and no grammatical errors. What's more, the extractive document summarization is easy for machine to generate summary so there are many models for extractive document summarization tasks.

The main purpose of abstractive document summarization is to understand the original document through natural language processing techniques, including shallow and deep understanding. Then based on the words in the original document as a basic unit, doing some operations like the word reorganization, replacement, synthesis and other operations, to generate the final summary. It is obvious that the abstractive document summarization can generate summary in a more human-like way than the extractive document summarization. At present, researchers are always using deep neural network to generate abstractive summary.

\section{Joint topic model and manifold ranking method for document summarization}
\subsection{Problem Statement}
Extractive document summarization is the common solution for document summarization tasks, especially for multi-document summarization tasks. For example, Wan et al. \shortcite{wan2007manifold} proposed a novel method to deal with the multi-document summarization task. The method can rank the importance of sentences by using manifold ranking. It mainly includes two steps: (1) Use manifold ranking to calculate the ranking scores for each sentence, where the score indicate the information richness of a sentence; (2) based on the manifold-ranking scores, it imposes a diversity of penalties to obtain the final ranking score of each sentence to reflect the richness of information and the novelty of information. However, the method of calculating the similarity between documents relies on the document representation method from word-to-document space. This method can only measure the co-occurrence of words between the surface words and cannot express the deep semantic of words. There are certain deficiencies.

In order to solve the above problems, we propose an document summarization model based on non-negative matrix factorization topic model and manifold ranking method. It not only uses the surface features of documents but also uses the semantic features to generate the better summary.
\subsection{Modeling: JTMMR}
In manifold ranking, you need to construct a weighted network by using data points. For extractive document summarization tasks, the sentences in the document can be considered as their data points. Similarly, for a topic-focused multi-document document summarization task, the topic description sentence is also one of the data points, and the data point has an indicating effect. Therefore, for a document summarization task, it can be defined as follows: Given a weighted network $G=(X, E)$, where $X={x_1,x_2,...,x_{N+1}}$ are the data points in the network and $E$ is the edge between points. $N$ is the number of sentences in the document and $N+1$ represents the total number of sentences after adding a topic description sentence to the sentence in the document. Therefore, by constructing a weighted matrix and updating equations of the manifold order, the ranking scores of each sentence can be obtained. We also define a vector $Y={y_1,y_2,...,y_{N+1}}$, in which $y_1=1$ because $x_1$ is the topic sentence and $y_i=0 (2 \geq i \geq N+1)$ for all the sentences in the
documents. We can get a loss function by joint non-negative topic model and manifold ranking method (JTMMR) as follow:
\begin{equation} \label{JTMMR}
\begin{array}{*{20}{c}}
\begin{array}{l}
{\rm{L(}}f,U,V{\rm{) = }}{\alpha _{{\rm{mr}}}}*{\sum\limits_{i,j = 1}^N {{W_{ij}}\left| {\frac{1}{{\sqrt {{D_{ii}}} }}{f_i} - \frac{1}{{\sqrt {{{\rm{D}}_{{\rm{jj}}}}} }}{f_j}} \right|} ^2} + (1 - {\alpha _{mr}})*
{\sum\limits_{i = 1}^N {\left| {{f_i} - {y_i}} \right|} ^2}\\\qquad\qquad\quad + \left\| {A - UV} \right\|_F^2 + \beta {\rm{*}}\left\| {{U^T}U - I} \right\|_F^2 + \lambda *tr({1^T}V1)\end{array}\\
{s.t.\left\{ {\begin{array}{*{20}{c}}
{{U_{i,j}} \ge 0}&{1 \le i \le M,1 \le j \le K}\\
{{V_{i,j}} \ge 0}&{1 \le i \le K,1 \le j \le N+1}
\end{array}} \right.}
\end{array}
\end{equation}
where $\beta$ is an orthogonal constraint hyper-parameter, the greater the value, the more independent between topics, and $\lambda$ is a sparse constraint hyper-parameter for $V$ matrix.

In order to joint the surface features of the original document and the semantic features between the documents, we define the formula for calculating the weighted matrix W in the manifold ranking as follows:
\begin{equation}
\begin{aligned}
\begin{array}{*{20}{c}}
{W_{ij}} = {\alpha _V}*\cos ({V_{*i}},{V_{*j}}){\rm{ + }}{\alpha _A}*\cos ({A_{*i}},{A_{*j}}) + {\alpha _{overlap}}*S{S_{i,j}}\\
{s.t.{\kern 1pt} {\kern 1pt} {\alpha _V} + {\alpha _A}{\rm{ + }}{\alpha _{overlap}}{\rm{ = 1}}}
\end{array}
\end{aligned}
\end{equation}
where $\cos ({V_{*i}},{V_{*j}})$ represent the cosine similarity of each document at the semantic level; $\cos ({A_{*i}},{A_{*j}})$ represent the cosine similarity of the surface feature level of the original document; the $SS$ matrix is the word unigram overlap matrix between the sentences, and it also belongs to the feature of the original document. Its definition is as follows:
\begin{equation}
\begin{aligned}
S{S_{i,j}} = \frac{{overlap({S_i},{S_j})}}{{{f_{len}}({S_i})}}
\end{aligned}
\end{equation}
By using three hyper-parameters $\alpha_V$, $\alpha_A$, $\alpha_{overlap}$, we can control the proportion of original document features and deep semantic features.
\subsection{Algorithm Derivations}
With respect to the optimization problem \ref{JTMMR}, it is nonconvex so there is not a global solution. However, we can seek for a local minimal for the practical applications. We can get the derivative of $L$ to $U$ is
\begin{equation}
\begin{aligned}
\frac{{\partial {{\rm{L}}_{tm}}(U,V)}}{{\partial U}} = 2UV{V^T} - 2A{V^T} + 4\beta U{U^T}U - 4\beta U
\end{aligned}
\end{equation}
the derivative of $L$ to $V$ is
\begin{equation}
\begin{aligned}
\frac{{\partial {L_{tm}}(U,V)}}{{\partial V}} = 2{U^T}UV - 2{U^T}A + \lambda {1_K}1_N^T
\end{aligned}
\end{equation}
Then we adopt the gradient descent method for each (ij)-th variable of $U$ and $V$ as below:
\begin{equation}
\begin{aligned}
{U_{m,k}} = {U_{m,k}}{\rm{*}}\frac{{{{(A{V^T} + 2\beta U)}_{m,k}}}}{{{{(UV{V^T} + 2\beta U{U^T}U)}_{m,k}}}}
\end{aligned}
\end{equation}
\begin{equation}
\begin{aligned}
{V_{k,n}} = {V_{k,n}}{\rm{*}}\frac{{{{({U^T}A)}_{k,n}}}}{{{{({U^T}UV + \frac{\lambda }{2}{1_K}1_N^T)}_{k,n}}}}
\end{aligned}
\end{equation}
Finally, we can use the JTMMR model to get the ranking score of each sentence. The corresponding algorithm for our proposed method is designed in Algo.~\ref{algo:JTMMR}:
\begin{algorithm}[t]
\caption{JTMMR}\label{algo:JTMMR}
\KwIn{Document $D$}
\KwOut{$U$, $V$, $f$}
\Begin
{
    Randomly initialized matrix $U$ and $V$\;
    initialized ranking score vector $f$ to all 0\;
    Optimize matrix $U$ and $V$\;
    Get matrix $SS$ according to $D$\;
    Get matrix $W$ according to $V$\;
    Use manifold ranking method to get ranking score $f$\;
    Return $U$, $V$, $f$\;
}
\end{algorithm}
\subsection{Summary Generation}
The quality of a summary is mainly reflected in three places: salience, redundancy, and fluency. Since we use the combination of the non-negative matrix factorization topic model and the manifold ranking method proposed in this paper, we can sort sentences according their saliency. In addition, the sentence we extract as the summary is the original sentence in the document, so the fluency of the summary can be guaranteed. It means that the main problem of the summary generation is the sentence redundancy. We use the MMR method similar to the paper \cite{wan2007manifold}. It is because the MMR method is easy to implement, and the computational complexity is not high.

The MMR algorithm we used to generate the summary is in Algo.~\ref{algo:MMR}.
\begin{algorithm}[t]
\caption{MMR}\label{algo:MMR}
\KwIn{$W$, $f$, $L$, $\omega$}
\KwOut{Summary $\Psi$}
\Begin
{
    Initialized $\Psi=\emptyset$, sentence $S={S_i|i=1,2,...,N}$\;
    W is normalized by $S={diag(W\bullet1)}^{-1}W \in R^{N*N}$\;
    \While{$len(\Psi) \le L$}{
        add the $S_{\max f_i}$ with the highest score to the summary $\Psi$, $\Psi {\rm{ = }}\Psi  \cup {S_{\max {f_i}}}$\;
        update the ranking score of sentence $f({S_i}) = f({S_i}) - \omega *{S_{i,\max {f_i}}}*f({S_{\max {f_i}}})$\;
    }
    Return $\Psi$\;
}
\end{algorithm}
\section{Joint lifelong topic model and manifold ranking method for document summarization}
\label{JOINT LIFELONG TOPIC MODEL AND MANIFOLD RANKING METHOD FOR DOCUMENT SUMMARIZATION}
\subsection{Lifelong Topic Model Based on Non-negative Matrix Factorization}
\subsubsection{Problem Statement}
The NMF topic model only focus on a single batch data and it cannot handle big data because of its time complexity. Moreover, each sentence is regarded as a document in the model for extractive document summarization task, while each sentence is often short. The short text problem will limit the performance of topic model \cite{yan2013learning} because the co-occurrence of words in the document is too few. It is the reason why the further improvement of the JTMMR model we proposed in Section 3 is limited.

In order to solve the above problems, we propose a lifelong topic model, which excavates higher quality implicit semantic features to achieve better document representation.
\subsubsection{Modeling: LTM}
\begin{figure}
\centering
\includegraphics[width=0.8\textwidth]{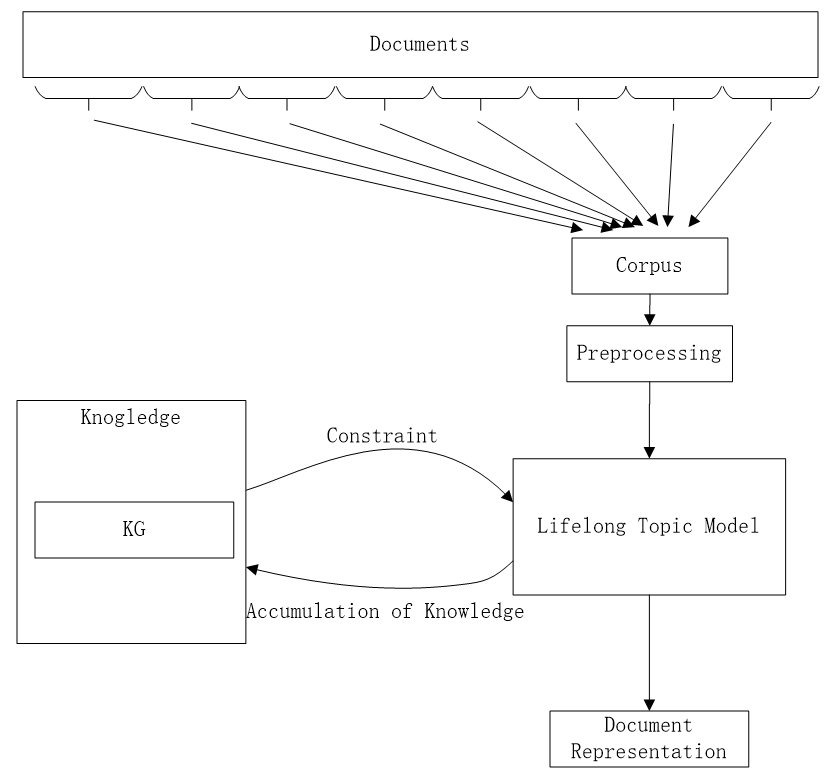}
\caption{Lifelong Topic Model}
\label{Lifelong Topic Model}
\end{figure}
Given a set of documents, which contains $N$ documents, and all the documents are word segmented to get a vocabulary with $M$ words. Then, we can form a document-to-word feature matrix $A$ according to the format of TFIDF value. To increase the orthogonal constraints of the $U$ matrix (word-to-topic) and the sparseness constraints of the $V$ matrix (topic-to-document) in the model, we can get:
\begin{equation}
\begin{aligned}
\begin{array}{*{20}{c}}
{\arg \min \{ \left\| {A - UV} \right\|_F^2 + \beta {\rm{*}}\left\| {{U^T}U - I} \right\|_F^2 + \lambda *tr({1^T}V1)\} }\\
{s.t.\left\{ {\begin{array}{*{20}{c}}
{{U_{i,j}} \ge 0}&{1 \le i \le M,1 \le j \le K}\\
{{V_{i,j}} \ge 0}&{1 \le i \le K,1 \le j \le N+1}
\end{array}} \right.}
\end{array}
\end{aligned}
\end{equation}
In order to transform the non-negative matrix factorization topic model into a lifelong topic model, we add the knowledge constraint and the constraints between document relationships to the model:
\begin{equation} \label{LTM}
\begin{aligned}
\begin{array}{*{20}{c}}
{\rm{L_{ltm}(}}U,V{\rm{) = }}\arg \min \{ \left\| {A - UV} \right\|_F^2 + \beta {\rm{*}}\left\| {{U^T}U - I} \right\|_F^2 
+ \lambda *tr({1^T}V1)\\{\rm{ + }}{\alpha _{ltm}}*tr({U^T}LU) + \gamma *tr(VZ{V^T})\} \\
{s.t.\left\{ \begin{array}{l}
\begin{array}{*{20}{c}}
{{U_{i,j}} \ge 0}&{1 \le i \le M,1 \le j \le K}\\
{{V_{i,j}} \ge 0}&{1 \le i \le K,1 \le j \le N+1}
\end{array}\\
\begin{array}{*{20}{c}}
{L = P - O}&{P = diag(O \cdot 1)}\\
{Z = T - Q}&{T = diag(Q \cdot 1)}
\end{array}
\end{array} \right.}
\end{array}
\end{aligned}
\end{equation}
where $O$ represents the knowledge. Here, we use is the number of co-occurrence word pairs in the subject. We use the top 10 keywords with the largest weight to represent each topic. Thus, the $O$ matrix is specifically defined as follows:
\begin{equation}
\begin{aligned}
\begin{array}{*{20}{c}}
{{O_{i,j}} = \frac{{\sum\limits_{previous{\kern 1pt} task} {\sum\limits_{topic} {\# {{({w_i},{w_j})}_{top10}}} } }}{{\sum\limits_{previous{\kern 1pt} task} {\sum\limits_{topic} {\sum\limits_{w \in \{ wordmap\} } {\# {{({w_i},w)}_{top10}}} } } }}}\\
{O \in {\mathbb{R}^{M*M}}}
\end{array}
\end{aligned}
\end{equation}
It means the lifelong topic model will store word pairs for the top 10 words of each topic into the knowledge in the past tasks. This kind of co-occurrence relationship of words can constrain the generation of $U$ matrix by the non-negative matrix factorization topic model, making the relationship between the generated topics-to-words more accurate.

$Q$ represents the constraints between document relationships:
\begin{equation}
\begin{aligned}
\begin{array}{*{20}{c}}
{{Q_{i,j}} = \frac{{sim(do{c_i},do{c_j})}}{{\sum\limits_{doc \in \{ docs\} } {sim(do{c_i},doc)} }}}\\
{Q \in {\mathbb{R}^{N*N}}}
\end{array}
\end{aligned}
\end{equation}
We can see that we have transformed the non-negative matrix factorization topic model to a lifelong topic model. The only difference between non-negative matrix factorization topic model and lifelong topic model is that we have added four constraints to the objective function. So the derivation is similar to the non-negative matrix factorization.
\subsubsection{Algorithm Derivations}
With respect to the optimization problem \ref{LTM}, it is nonconvex so there is not a global solution. However, we can seek for a local minimal for the practical applications. We can get the derivative of $L$ to $U$ is
\begin{equation}
\begin{aligned}
\frac{{\partial {{\rm{L}}_{ltm}}(U,V)}}{{\partial U}} = 2UV{V^T} - 2A{V^T} + 2{\alpha _{ltm}}PU - 2{\alpha _{ltm}}OU + 4\beta U{U^T}U - 4\beta U
\end{aligned}
\end{equation}
the derivative of $L$ to $V$ is
\begin{equation}
\begin{aligned}
\frac{{\partial {L_{ltm}}(U,V)}}{{\partial V}} = 2{U^T}UV - 2{U^T}A + \lambda {1_K}1_N^T
\end{aligned}
\end{equation}
Then we adopt the gradient descent method for each (ij)-th variable of $U$ and $V$ as below:
\begin{equation}
\begin{aligned}
{U_{m,k}} = {U_{m,k}}{\rm{*}}\frac{{{{(A{V^T} + {\alpha _{ltm}}OU + 2\beta U)}_{m,k}}}}{{{{(UV{V^T} + {\alpha _{ltm}}PU + 2\beta U{U^T}U)}_{m,k}}}}
\end{aligned}
\end{equation}
\begin{equation}
\begin{aligned}
{V_{k,n}} = {V_{k,n}}{\rm{*}}\frac{{{{({U^T}A)}_{k,n}}}}{{{{({U^T}UV + \frac{\lambda }{2}{1_K}1_N^T)}_{k,n}}}}
\end{aligned}
\end{equation}
Finally, the algorithm of lifelong topic model is in Algo.~\ref{algo:LTM}.
\begin{algorithm}[t]
\caption{Lifelong Topic Model}\label{algo:LTM}
\KwIn{Documents $D={D_1,D_2,...,D_N}$, Relationship between Documents $Q={Q_1,Q_2,...,Q_N}$}
\KwOut{$U$, $V$, $KG$}
\Begin
{
    initialized knowledge matrix $KG \in R^{M*M}$ to all 0\;
    \For{$i=1\ to \ N$}{
        preprocessing document $D_i$\;
        generate matrix $A_i$ according to TFIDF\;
        randomly initialized matrix $U$ and $V$\;
        \While{not convergence}{
            update $U$\;
            update $V$\;
        }
        $KG=recordKnowledge(U)$\;
    }
    Return $U$, $V$, $KG$\;
}
\end{algorithm}
\subsection{Joint Lifelong Topic Model and Manifold Ranking for Document Summarization}
\subsubsection{Problem Statement}
Similar to the JTMMR model we proposed in Section 3, we use the lifelong topic model to replace the non-negative matrix factorization topic model in the JTMMR model, decompose the original document, and then use the original features of the document and the semantic features to better represent documents. The result of the document representation is then used to generate the saliency of the sentence by manifold ranking. After obtaining the ranking score of sentences, we will adjust the lifelong topic model. Iterate lifelong topic model and manifold ranking until convergence, and finally the local optimal solution could be achieved.
\subsubsection{Modeling: JLTMMR}
We use the lifelong topic model to replace the non-negative matrix factorization topic model in the JTMMR model, and then combine the lifelong topic model and manifold ranking methods (JLTMMR) to obtain the following optimized equation:
\begin{equation}
\begin{aligned}
\begin{array}{*{20}{c}}
\begin{array}{l}
{\rm{L(}}f,U,V{\rm{) = }}{\alpha _{{\rm{mr}}}}*{\sum\limits_{i,j = 1}^N {{W_{ij}}\left| {\frac{1}{{\sqrt {{D_{ii}}} }}{f_i} - \frac{1}{{\sqrt {{{\rm{D}}_{{\rm{jj}}}}} }}{f_j}} \right|} ^2} + (1 - {\alpha _{mr}})*{\sum\limits_{i = 1}^N {\left| {{f_i} - {y_i}} \right|} ^2}\\
\qquad\qquad\qquad + \left\| {A - UV} \right\|_F^2 + \beta {\rm{*}}\left\| {{U^T}U - I} \right\|_F^2 + \lambda *tr({1^T}V1)\\
\qquad\qquad\qquad {\rm{ + }}{\alpha _{ltm}}*tr({U^T}LU) + \gamma *tr(VZ{V^T})
\end{array}\\
{s.t.\left\{ \begin{array}{l}
\begin{array}{*{20}{c}}
{{U_{i,j}} \ge 0}&{1 \le i \le M,1 \le j \le K}\\
{{V_{i,j}} \ge 0}&{1 \le i \le K,1 \le j \le N+1}
\end{array}\\
\begin{array}{*{20}{c}}
{L = P - O}&{P = diag(O \cdot 1)}\\
{Z = T - Q}&{T = diag(Q \cdot 1)}
\end{array}
\end{array} \right.}
\end{array}
\end{aligned}
\end{equation}
where $Q$ represents priori condition of the documents relationship:
\begin{equation}
\begin{aligned}
\begin{array}{*{20}{c}}
{{Q_{_{ij}}} = 1 - \left| {{f_i} - {f_j}} \right|}&{f \stackrel{map\ to}{\longrightarrow} [0,1]}
\end{array}
\end{aligned}
\end{equation}
In addition, in order to enhance the role of ranking score for lifelong topic model, a weighted matrix $R$ is added to the model.
\begin{equation}
\begin{aligned}
\begin{array}{*{20}{c}}
{{R_{ij}} = {e^{{f_j}}}}&{f \stackrel{map\ to}{\longrightarrow} [0,1]}
\end{array}
\end{aligned}
\end{equation}
Overall, the final optimized equation of JLTMMR is as follows:
\begin{equation} \label{JLTMMR}
\begin{aligned}
\begin{array}{*{20}{c}}
\begin{array}{l}
{\rm{L(}}f,U,V{\rm{) = }}{\alpha _{{\rm{mr}}}}*{\sum\limits_{i,j = 1}^N {{W_{ij}}\left| {\frac{1}{{\sqrt {{D_{ii}}} }}{f_i} - \frac{1}{{\sqrt {{{\rm{D}}_{{\rm{jj}}}}} }}{f_j}} \right|} ^2} + (1 - {\alpha _{mr}})*{\sum\limits_{i = 1}^N {\left| {{f_i} - {y_i}} \right|} ^2}\\
\qquad\qquad\qquad + R \odot \left\| {A - UV} \right\|_F^2 + \beta {\rm{*}}\left\| {{U^T}U - I} \right\|_F^2 + \lambda *tr({1^T}V1)\\
\qquad\qquad\qquad {\rm{ + }}{\alpha _{ltm}}*tr({U^T}LU) + \gamma *tr(VZ{V^T})
\end{array}\\
{s.t.\left\{ \begin{array}{l}
\begin{array}{*{20}{c}}
{{U_{i,j}} \ge 0}&{1 \le i \le M,1 \le j \le K}\\
{{V_{i,j}} \ge 0}&{1 \le i \le K,1 \le j \le N+1}
\end{array}\\
\begin{array}{*{20}{c}}
{L = P - O}&{P = diag(O \cdot 1)}\\
{Z = T - Q}&{T = diag(Q \cdot 1)}
\end{array}\\
\begin{array}{*{100}{c}}
{{y_i} = 1}&{{X_i \quad is \quad topic \quad sentence }}&{}\\
{{y_i} = 0}&{{X_i \quad is \quad not \quad topic \quad sentence}}
\end{array}\\
\begin{array}{*{20}{c}}
{W_{ij}} = {\alpha _V}*\cos ({V_{*i}},{V_{*j}}){\rm{ + }}{\alpha _A}*\\ \cos ({A_{*i}},{A_{*j}}) + {\alpha _{overlap}}*S{S_{i,j}}
\end{array}\\
\begin{array}{*{20}{c}}
{{\alpha _V} + {\alpha _A}{\rm{ + }}{\alpha _{overlap}}{\rm{ = 1}}}
\end{array}\\
\begin{array}{*{20}{c}}
{{Q_{_{ij}}} = 1 - \left| {{f_i} - {f_j}} \right|}&{f \stackrel{map\ to}{\longrightarrow} [0,1]}
\end{array}\\
\begin{array}{*{20}{c}}
{{R_{ij}} = {e^{{f_j}}}}&{f \stackrel{map\ to}{\longrightarrow} [0,1]}
\end{array}
\end{array} \right.}
\end{array}
\end{aligned}
\end{equation}

\subsubsection{Algorithm Derivations}
With respect to the optimization problem \ref{JLTMMR}, it is nonconvex so there is not a global solution. However, we can seek for a local minimal for the practical applications. We can get the derivative of $L$ to $U$ is
\begin{equation}
\begin{aligned}
\begin{array}{l}
\frac{{\partial {{\rm{L}}_{jltmmr}}(U,V)}}{{\partial U}} = 2U(V \odot R'){(V \odot R')^T} - 2(A \odot R'){(V \odot R')^T}\\
\qquad\qquad\qquad\qquad+ 2{\alpha _{ltm}}PU - 2{\alpha _{ltm}}OU + 4\beta U{U^T}U - 4\beta U
\end{array}
\end{aligned}
\end{equation}
the derivative of $L$ to $V$ is
\begin{equation}
\begin{aligned}
\frac{{\partial {L_{jltmmr}}(U,V)}}{{\partial V}} = 2(({U^T}U(V \odot R')) \odot R') - 2(({U^T}(A \odot R')) \odot R') + \lambda {1_K}1_N^T
\end{aligned}
\end{equation}
Then we adopt the gradient descent method for each (ij)-th variable of $U$ and $V$ as below:
\begin{equation}
\begin{aligned}
{U_{m,k}} = {U_{m,k}}{\rm{*}}\frac{{{{((A \odot R'){{(V \odot R')}^T} + {\alpha _{ltm}}OU + 2\beta U)}_{m,k}}}}{{{{(U(V \odot R'){{(V \odot R')}^T} + {\alpha _{ltm}}PU + 2\beta U{U^T}U)}_{m,k}}}}
\end{aligned}
\end{equation}
\begin{equation}
\begin{aligned}
{V_{k,n}} = {V_{k,n}}{\rm{*}}\frac{{{{((({U^T}(A \odot R')) \odot R'))}_{k,n}}}}{{{{((({U^T}U(V \odot R')) \odot R') + \frac{\lambda }{2}{1_K}1_N^T)}_{k,n}}}}
\end{aligned}
\end{equation}
Finally, we can use the JLTMMR model to get the ranking score of each sentence. The algorithm is in Algo.~\ref{algo:JLTMMR}.
\begin{algorithm}[t]
\caption{JLTMMR}\label{algo:JLTMMR}
\KwIn{Document $D$}
\KwOut{$U$, $V$, $f$, $KG$}
\Begin
{
    initialized knowledge matrix $KG \in R^{M*M}$ to all 0\;
    initialized ranking score vector $f$ to all 0\;
    \While{not convergence}{
        get matrix $Q$ and $R$ according to $f$\;
        get matrix $O$ according to $KG$\;
        Optimize matrix $U$ and $V$\;
        $KG=recordKnowledge(U)$\;
        get matrix $W$ according to $V$\;
        Use manifold ranking method to get ranking score $f$\;
    }
    Return $U$, $V$, $f$, $KG$\;
}
\end{algorithm}
\subsubsection{Summary Generation}
Here we also use the MMR method in the JTMMR model to generate the summary. For the calculation of the sentence saliency, we use the statistical features (SF) similar to the paper \cite{ren2017leveraging} except for the sentence scores generated in the JLTMMR model, as shown in Table \ref{Statistical Features}.
\begin{table}
  \caption{Statistical Features}
  \label{Statistical Features}
  \begin{minipage}{\textwidth}
    \begin{tabular}{p{6.5cm}p{6cm}}
    \hline\hline
     Statistical Features & Description\\
    \hline
     ${f_{tf}}({S_i}) = \frac{{\sum\limits_{w \in {S_i}} {TF(w)} }}{{{f_{len}}({S_i})}}$ & Average term frequency. $TF(w)$ is the term frequency of word $w$ \\
     \noalign{\vspace {.5cm}}
     ${f_{sim\_TS}}({S_t},{S_i}) = \cos (emb({S_t}),emb({S_i}))$ & The similarity between topic sentence $S_t$ and sentence $S_i$ \\
     \noalign{\vspace {.5cm}}
     ${f_{overlap\_{S_t}}}({S_t},{S_i}) = \frac{{overlap({S_t},{S_i})}}{{{f_{len}}({S_t})}}$ & The unigram overlap between topic sentence $S_t$ and sentence $S_i$ relative to $S_t$ \\
     \noalign{\vspace {.5cm}}
     ${f_{overlap\_{S_i}}}({S_t},{S_i}) = \frac{{overlap({S_t},{S_i})}}{{{f_{len}}({S_i})}}$ & The unigram overlap between topic sentence $S_t$ and sentence $S_i$ relative to $S_i$ \\
     \noalign{\vspace {.5cm}}
     ${f_{pos}}({S_i})$ & Position of $S_i$ in its document\\
    \hline\hline
    \end{tabular}
    \vspace{-2\baselineskip}
  \end{minipage}
\end{table}
\\
\\
\\
The final ranking score of JLTMMR+SF is as follows:
\begin{equation}
\begin{aligned}
\begin{array}{l}
f({S_i}) = {f_{jlmlmr}}({S_i}) + {\mu _1}{\rm{*}}{f_{tf}}({S_i}) + {\mu _2}{\rm{*}}{f_{sim\_TS}}({S_t},{S_i}) \\
\qquad\qquad + {\mu _3}{\rm{*}}{f_{overlap\_{S_t}}}({S_t},{S_i}) + {\mu _4}{\rm{*}}{f_{overlap\_{S_i}}}({S_t},{S_i}) + {\mu _5}{\rm{*}}{f_{pos}}({S_i})
\end{array}
\end{aligned}
\end{equation}
\section{Experiments}
\subsection{Dataset}
We conduct experiments on four datasets of DUC 2005\footnote{http://duc.nist.gov/duc2005/tasks.html}, DUC 2006\footnote{http://duc.nist.gov/duc2006/tasks.html}, DUC 2007\footnote{http://duc.nist.gov/duc2007/tasks.html} main task and NLPCC 2015\footnote{http://tcci.ccf.org.cn/conference/2015/index.html}. The statistics of the four datasets is listed in Table~\ref{The statistics of the four datasets}.
\begin{table}
  \caption{The statistics of the four datasets. MDS and SDS indicate multi-document summarization and single document summarization, respectively. Each document is a single topic for single document summarization task.}
  \label{The statistics of the four datasets}
  \begin{minipage}{\textwidth}
    \begin{tabular}{lllll}
    \hline\hline
      & DUC 2005 & DUC 2006 & DUC 2007 & NLPCC 2015 \\
    \hline
     Task & Only one & Only one & Main task & Task 4 \\
     \noalign{\vspace {.5cm}}
     Style & MDS & MDS & MDS & SDS \\
     \noalign{\vspace {.5cm}}
     Language & English & English & English & Chinese \\
     \noalign{\vspace {.5cm}}
     Topic Number & 50 & 50 & 45 & 250 \\
     \noalign{\vspace {.5cm}}
     Document Number of \\ Each Topic & 32 & 25 & 25 & 1 \\
     \noalign{\vspace {.5cm}}
     Document source & TREC & AQUAINT & AQUAINT & Sina Weibo \\
     \noalign{\vspace {.5cm}}
     Length of abstract & 250 words & 250 words & 250 words & 140 characters \\
    \hline\hline
    \end{tabular}
    \vspace{-2\baselineskip}
  \end{minipage}
\end{table}

The DUC 2005, DUC 2006 and DUC 2007 datasets are for query-focused multi-document summarization. The documents are from the news domain and grouped into thematic clusters. There is a short topic statement in each cluster and the topic statement could be in the form of a question or set of related questions and could include background information that the assessor thought would help clarify his/her information need. The NLPCC 2015 dataset is for single document summarization. The document is from Chinese news articles which posted on Sina Weibo.
\subsection{Data processing}
For each document cluster, we concatenate all the documents and split them into sentences by using the tool provided in the DUC 2003 dataset\footnote{http://duc.nist.gov/duc2003/tasks.html}. After that, we use the data processing step which is similar to the paper \cite{wan2007manifold}. Firstly, we remove the dialog sentences and stop words in each sentence. Secondly, for English datasets, the remaining words in each sentence were stemmed by using the Porter's stemmer \cite{van1980new} and for Chinese datasets, we use THULAC (THU Lexical Analyzer for Chinese) \cite{sun2016thulac} for Chinese word segmentation. The statistics of four datasets after data processing is listed in Table~\ref{The statistics of the four datasets after processing}.
\begin{table}
  \caption{The statistics of the four datasets after processing.}
  \label{The statistics of the four datasets after processing}
  \begin{minipage}{\textwidth}
    \begin{tabular}{p{3.8cm}p{1.6cm}p{1.6cm}p{1.6cm}p{2.5cm}}
    \hline\hline
      {} & {DUC 2005} & {DUC 2006} & {DUC 2007} & {NLPCC 2015} \\
    \hline
     Average Sentence Number In Each Topic & 799.92 & 598.78 & 454.69 & 45.21 \\
     \noalign{\vspace {.5cm}}
      Average Words Number In Each Sentence & 13.48 & 13.46 & 13.28 & 9.41 \\
     \noalign{\vspace {.5cm}}
      Size of Wordmap & 21916 & 17797 & 14944 & 14368 \\
    \hline\hline
    \end{tabular}
    \vspace{-2\baselineskip}
  \end{minipage}
\end{table}
\begin{table}
  \caption{Methods considered for comparison}
  \begin{minipage}{\textwidth}
    \begin{tabular}{p{2.1cm}p{5.6cm}p{3.95cm}}
    \hline\hline
       {Acronym} & {Gloss} & {Reference} \\
    \hline
     SingleMR &  & ~\ref{JOINT LIFELONG TOPIC MODEL AND MANIFOLD RANKING METHOD FOR DOCUMENT SUMMARIZATION} \\
 JLTMMR & The proposed model in this paper & ~\ref{JOINT LIFELONG TOPIC MODEL AND MANIFOLD RANKING METHOD FOR DOCUMENT SUMMARIZATION} \\
 JLTMMR+SF & Combination of JLTMMR and SF & ~\ref{JOINT LIFELONG TOPIC MODEL AND MANIFOLD RANKING METHOD FOR DOCUMENT SUMMARIZATION} \\
     \noalign{\vspace {.5cm}}
      \multicolumn{2}{l}{\emph{\textbf{Traditional methods}}} &  \\
 Random & Select sentences randomly for each topic & - \\
 Lead & Take the first sentences one by one in the recent document & - \\
 Kmeans+NMF & Combination of Kmeans and NMF based method & \cite{park2007multi} \\
    \noalign{\vspace {.5cm}}
     \multicolumn{2}{l}{\emph{\textbf{Neural network based methods}}} &  \\
 MV-CNN & Multi-view convolutional neural networks method & \cite{zhang2017multiview} \\
 QODE & Query-oriented deep extraction method & \cite{liu2012query} \\
 AttSum & Neural attention based method & \cite{cao2016attsum} \\
 VAEs-A & Variational auto-encoders based method & \cite{li2017salience} \\
 C-Attention & Cascaded attention based unsupervised model & \cite{li2017cascaded} \\
     \noalign{\vspace {.5cm}}
       \multicolumn{2}{l}{\emph{\textbf{Manifold ranking based methods}}} &  \\
 MultiMR & Multi-modality manifold-ranking method & \cite{wan2009graph} \\
 JMFMR &  Joint matrix factorization and manifold-ranking method & \cite{tan2015joint} \\
    \hline\hline
    \end{tabular}
    \vspace{-2\baselineskip}
  \end{minipage}
\end{table}

\subsection{Evaluation Protocols}

\quad(i) Topic Coherence

Coherence \cite{newman2010automatic,lau2014machine} was firstly proposed by David Mimno to evaluate the topics’ performance. Its formulation is as follows:
\begin{equation}
\begin{aligned}
Coherence(t;{V^{(t)}}) = \sum\limits_{m = 2}^M {\sum\limits_{l = 1}^{m - 1} {\log \frac{{\# (v_m^{(t)},v_l^{(t)}) + 1}}{{\# (v_l^{(t)})}}} }
\end{aligned}
\end{equation}
where $\# (v_m^{(t)},v_l^{(t)})$ denotes the number of documents that contains the two terms of a latent topic, while $\# (v_l^{(t)})$ marks the number of documents where the $l^{th}$ term appears. The larger value of coherence it has, the better performance of the topic.

(ii) ROUGE

The commonly used evaluation method of document summarization is ROUGE (Recall-Oriented Understudy for Gisting Evaluation) \cite{lin2004rouge}, which is also used by the DUC. ROUGE can automatically determine the quality of machine-generated summary by comparing the manually-edited summary and machine-generated summary. The ROUGE method measures the quality of the summary by measuring the number of overlapping units between the machine-generated summary and the manually-edited summary, such as N-grams, word sequences, and word pairs. ROUGE-N is used to calculate the degree of overlap by using the N-gram. The formula is as follows:
\begin{equation}
\begin{aligned}
ROUGE - N{\rm{ = }}\frac{{\sum\limits_{S \in \{ {\mathop{\rm Re}\nolimits} fSummary\} } {\sum\limits_{gra{m_n} \in S} {Coun{t_{match}}(gra{m_n})} } }}{{\sum\limits_{S \in \{ {\mathop{\rm Re}\nolimits} fSummary\} } {\sum\limits_{gra{m_n} \in S} {Count(gra{m_n})} } }}
\end{aligned}
\end{equation}
where $n$ represents the use of N-gram, $Coun{t_{match}}(gra{m_n})$ represents the overlap of N-grams between manually-edited summary and machine-generated summary, $Count(gra{m_n})$ represents the total number of N-grams. In addition, ROUGE also has ROUGE-W and ROUGE-SU for further measure.

(iii) Human Evaluation

Here we use the same human evaluation method as the DUC 2005 \cite{dang2005overview}. We mainly use five semantic quality evaluation criteria: 1) grammatically, 2) non-redundancy, 3) referential clarity, 4) focus, 5) structure and coherence. The range of human evaluation score is from 1 to 5, the larger value is better.
\subsection{Experiment Result}
The experiment result of ROUGE is shown in Table \ref{ROUGE Result} and Table \ref{NLPCC ROUGE Result}. The experiment result of human evaluation is shown in Table \ref{Human Evaluation Result}.
\subsubsection{DUC 2005 dataset}
It can be seen that the JLTMMR model has a lot of improvement over the SingleMR which only uses the original features of the sentence, and the JTMMR model which does not use the lifelong topic model. The JLTMMR+SF model obtained by adding statistical features has greatly improved compared to the JLTMMR model. This also shows that statistical features do play a big role in the generation of document summaries. The effect of the JLTMMR+SF model was better than that of all the baseline models including the model based on deep learning of the last 2 years. In the human evaluation results, the gap between them is more obvious.
\subsubsection{DUC 2006 dataset}
It can be seen that the JLTMMR+SF model achieves the best result except on ROUGE-2, which is less than the AttSum model. It proves that our model with original document features, semantic features and statistical features has a lot of advantages.
\subsubsection{DUC 2007 dataset}
In DUC 2007 dataset, our JLTMMR+SF model is slightly inferior to the AttSum model. The AttSum model uses a framework of deep learning. It uses CNN+Attention method to learn the distributed representation of sentences and documents. Although the JLTMMR+SF model proposed by us cannot achieve the best result, it also achieves comparable results by comparing with other baseline models.
\begin{table}[H]
  \caption{Rouge results of multi-document summarization on DUC 2005, DUC 2006, DUC 2007 datasets.}
  \label{ROUGE Result}
  \begin{minipage}{\textwidth}
    \begin{tabular}{llllll}
    \hline\hline
     {} & {System} & {Rouge-1} & {Rouge-2} & {Rouge-W} & {Rouge-SU} \\
    \hline
     \multirow{10}*{DUC 2005} & {Random} & {0.3152} & {0.04236} & {0.10809} & {0.09651} \\
 & {Lead} & {0.32098} & {0.046} & {0.1114} & {0.10173} \\
 & {Kmeans+NMF} & {0.33364} & {0.05306} & {0.11641} & {0.10804} \\
 & {QODE} & {0.3751} & {0.0775} & {-} & {0.1341} \\
 & {AttSum} & {0.3701} & {0.0699} & {-} & {-} \\
 & {MultiMR} & {0.37183} & {0.06761} & {0.12927} & {-} \\
 & {SingleMR} & {0.36347} & {0.06035} & {0.12295} & {0.11705} \\
 & \emph{\textbf{JTMMR}} & \emph{0.37158} & \emph{0.06758} & \emph{0.12747} & \emph{0.12519} \\
 & \emph{\textbf{JLTMMR}} & \emph{0.37410} & \emph{0.06953} & \emph{0.12846} & \emph{0.12679} \\
 & \emph{\textbf{JLTMMR+SF}} & \emph{\textbf{0.38401}} & \emph{\textbf{0.08035}} & \emph{\textbf{0.13274}} & \emph{\textbf{0.13891}} \\
 \noalign{\vspace {.5cm}}
     \multirow{13}*{DUC 2006} & {Random} & {0.35352} & {0.05381} & {0.12135} & {0.111} \\
 & {Lead} & {0.3396} & {0.05412} & {0.11748} & {0.10961} \\
 & {Kmeans+NMF} & {0.35856} & {0.06468} & {0.12493} & {0.12123} \\
 & {MV-CNN} & {0.3865} & {0.0791} & {-} & {0.1409} \\
 & {QODE} & {0.4015} & {0.0928} & {-} & {0.1479} \\
 & {AttSum} & {0.409} & \textbf{0.094} & {-} & {-} \\
 & {VAEs-A} & {0.396} & {0.089} & {-} & {0.143} \\
 & {C-Attention} & {0.393} & {0.087} & {-} & {0.141} \\
 & {MultiMR} & {0.40306} & {0.08508} & {0.13997} & {-} \\
     & {SingleMR} & {0.39934} & {0.07502} & {0.13445} & {0.13404} \\
 & \emph{\textbf{JTMMR}} & \emph{0.40014} & \emph{0.08160} & \emph{0.13652} & \emph{0.14062} \\
 & \emph{\textbf{JLTMMR}} & \emph{0.4043} & \emph{0.08126} & \emph{0.13774} & \emph{0.14032} \\
 & \emph{\textbf{JLTMMR+SF}} & \emph{\textbf{0.41045}} & \emph{0.08926} & \emph{\textbf{0.14214}} & \emph{\textbf{0.14798}} \\
     \noalign{\vspace {.5cm}}
    \multirow{13}*{DUC 2007} & {Random} & {0.36896} & {0.06654} & {0.125} & {0.12312} \\
 & {Lead} & {0.35461} & {0.06639} & {0.1219} & {0.11926} \\
 & {Kmeans+NMF} & {0.37076} & {0.07194} & {0.12786} & {0.12814} \\
 & {MV-CNN} & {0.4092} & {0.0911} & {-} & {0.1534} \\
 & {QODE} & {0.4295} & \textbf{0.1163} & {-} & \textbf{0.1685} \\
 & {AttSum} & \textbf{0.4392} & {0.1155} & {-} & {-} \\
 & {VAEs-A} & {0.421} & {0.11} & {-} & {0.164} \\
 & {C-Attention} & {0.423} & {0.107} & {-} & {0.161} \\
 & {MultiMR} & {0.42041} & {0.10302} & {0.14595} & {-} \\
     & {SingleMR} & {0.41422} & {0.09052} & {0.13984} & {0.14589} \\
 & \emph{\textbf{JTMMR}} & \emph{0.42717} & \emph{0.10181} & \emph{0.1458} & \emph{0.15761} \\
 & \emph{\textbf{JLTMMR}} & \emph{0.43349} & \emph{0.10375} & \emph{0.14786} & \emph{0.16002} \\
 & \emph{\textbf{JLTMMR+SF}} & \emph{0.43734} & \emph{0.10439} & \emph{\textbf{0.1496}} & \emph{0.1625} \\
    \hline\hline
    \end{tabular}
    \vspace{-2\baselineskip}
  \end{minipage}
\end{table}

\subsubsection{NLPCC 2015 dataset}
In NLPCC 2015 dataset, we can see that JLTMMR+SF only achieve the third place results on ROUGE-2. It is because our model is mainly designed for multi-document summarization task while NLPCC 2015 dataset is a single-document summarization task. In addition, according to the description of the best result \cite{liu2015weibo}, it can be found that their model is particularly simple. Their model can get a very good result by weighting four statistical features, including word frequency, sentence position, sentence length, sentence and title similarity. This shows that the statistical characteristics is very useful for the document summarization task, especially for single document summarization task. It also shows that our JLTMMR model can be greatly improved after adding statistical features.

Based on the above analysis of the results on each dataset, it can be clearly seen that the model JLTMMR we proposed can improve the performance compared with the JTMMR without using the lifelong learning method. At the same time, our JLTMMR+SF model also has certain advantages over the deep learning model proposed by other scholars in recent years.

\begin{table}[H]
  \caption{Rouge results of single-document summarization on NLPCC 2015 datasets.}
  \label{NLPCC ROUGE Result}
  \begin{minipage}{\textwidth}
    \begin{tabular}{llllll}
    \hline\hline
     {System} & {Rouge-1} & {Rouge-2} & {Rouge-3} & {Rouge-4} & {Rouge-SU4} \\
    \hline
    {Random} & {0.37234} & {0.20346} & {0.14447} & {0.11611} & {0.19139} \\
 {Lead} & {0.40779} & {0.25251} & {0.19033} & {0.15688} & {0.23269} \\
 {Kmeans+NMF} & {0.40435} & {0.24911} & {0.18783} & {0.15531} & {0.23196} \\
 {CCNUTextMiner} & {0.44166} & {0.28354} & {0.21853} & {0.18420} & {0.26633} \\
 {CIST-SUMM} & {0.47368} & {0.32354} & {0.26122} & {0.22869} & {0.30737} \\
 {DluflNLP} & {0.42166} & {0.26136} & {0.1958} & {0.16211} & {0.24464} \\
 {FLCTest4} & {0.48773} & {0.34913} & {0.28898} & {0.25574} & {0.33162} \\
 {NLP@WUST} & \textbf{0.52422} & \textbf{0.40035} & \textbf{0.34483} & \textbf{0.31234} & \textbf{0.38318} \\
 {USC1} & {0.41572} & {0.25132} & {0.1865} & {0.15548} & {0.23756} \\
 {zutnlptest4} & {0.39945} & {0.24392} & {0.18361} & {0.15297} & {0.23018} \\
 {YTSC-run} & {0.46747} & {0.32056} & {0.25901} & {0.22655} & {0.30448} \\
 {zzubylong} & {0.36854} & {0.22465} & {0.15649} & {0.12352} & {0.2} \\
 \noalign{\vspace {.5cm}}
 {SingleMR} & {0.39409} & {0.21907} & {0.15316} & {0.12284} & {0.20343} \\
 \emph{\textbf{JTMMR}} & \emph{0.39853} & \emph{0.22380} & \emph{0.15705} & \emph{0.12597} & \emph{0.20658} \\
 \emph{\textbf{JLTMMR}} & \emph{0.40015} & \emph{0.23146} & \emph{0.16661} & \emph{0.13500} & \emph{0.21508} \\
 \emph{\textbf{JLTMMR+SF}} & \emph{0.46599} & \emph{0.32769} & \emph{0.26779} & \emph{0.23355} & \emph{0.30867} \\
    \hline\hline
    \end{tabular}
    \vspace{-2\baselineskip}
  \end{minipage}
\end{table}

\begin{table}[H]
  \caption{Human Evaluation results of multi-document summarization on DUC 2005, DUC 2006, DUC 2007 and single-document summarization on NLPCC 2015 datasets.}
  \label{Human Evaluation Result}
  \begin{minipage}{\textwidth}
    \begin{tabular}{p{0.9cm}p{2.25cm}p{1.3cm}p{1.5cm}p{1.5cm}lp{1.9cm}}
    \hline\hline
 {} & {System} & {Gramma-ticality} & {Non-redundancy} & {Referential clarity} & {Focus} & {Structure and \qquad\qquad Coherence} \\
 \hline
 \multirow{5}{0.5cm}{DUC 2005} & {Lead} & {2.425} & {2.55} & {2.35} & {2.15} & {2.275} \\
 & {Kmeans+NMF} & {2.75} & {2.925} & {2.7} & {2.725} & {2.325} \\
 & {SingleMR} & {3.25} & {3.325} & {3.575} & {3.325} & {3.15} \\
 & \emph{\textbf{JTMMR}} & \emph{3.5} & \emph{3.65} & \emph{3.7} & \emph{3.6} & \emph{3.45} \\
 & \emph{\textbf{JLTMMR+SF}} & \emph{\textbf{3.9}} & \emph{\textbf{4.025}} & \emph{\textbf{3.95}} & \emph{\textbf{4.2}} & \emph{\textbf{3.725}} \\
 \noalign{\vspace {.5cm}}
 \multirow{5}{0.5cm}{DUC 2006} & {Lead} & {2.875} & {2.825} & {2.425} & {2.3} & {2.475} \\
 & {Kmeans+NMF} & {2.85} & {3.075} & {2.95} & {2.8} & {2.575} \\
 & {SingleMR} & {3.575} & {3.925} & {3.35} & {3.275} & {3.05} \\
 & \emph{\textbf{JTMMR}} & \emph{3.65} & \emph{4.025} & \emph{3.55} & \emph{3.425} & \emph{3.2} \\
 & \emph{\textbf{JLTMMR+SF}} & \emph{\textbf{3.95}} & \emph{\textbf{4.25}} & \emph{\textbf{3.8}} & \emph{\textbf{3.975}} & \emph{\textbf{3.35}} \\
 \noalign{\vspace {.5cm}}
 \multirow{5}{0.5cm}{DUC 2007} & {Lead} & {3.075} & {3.1} & {2.675} & {2.375} & {2.725} \\
 & {Kmeans+NMF} & {3.225} & {3.25} & {3.15} & {2.875} & {2.5} \\
 & {SingleMR} & {3.625} & {4} & {3.375} & {3.275} & {3.025} \\
 & \emph{\textbf{JTMMR}} & \emph{3.75} & \emph{4.025} & \emph{3.5} & \emph{3.475} & \emph{3.075} \\
 & \emph{\textbf{JLTMMR+SF}} & \emph{\textbf{3.975}} & \emph{\textbf{4.1}} & \emph{\textbf{3.725}} & \emph{\textbf{3.85}} & \emph{\textbf{3.175}} \\
 \noalign{\vspace {.5cm}}
 \multirow{5}{0.5cm}{NLPCC 2015} & {Lead} & {2.8} & {2.9} & {2.775} & {2.8} & {2.575} \\
 & {Kmeans+NMF} & {3.15} & {2.85} & {2.9} & {3.125} & {2.7} \\
 & {SingleMR} & {3.475} & {3.8} & {3.35} & {3.75} & {2.975} \\
 & \emph{\textbf{JTMMR}} & \emph{3.625} & \emph{3.95} & \emph{3.55} & \emph{3.875} & \emph{3.125} \\
 & \emph{\textbf{JLTMMR+SF}} & \emph{\textbf{3.95}} & \emph{\textbf{4.125}} & \emph{\textbf{3.825}} & \emph{\textbf{4.025}} & \emph{\textbf{3.5}} \\
    \hline\hline
    \end{tabular}
    \vspace{-2\baselineskip}
  \end{minipage}
\end{table}

\subsubsection{Example}
\begin{figure}[H]
\centering
\includegraphics[width=1\textwidth]{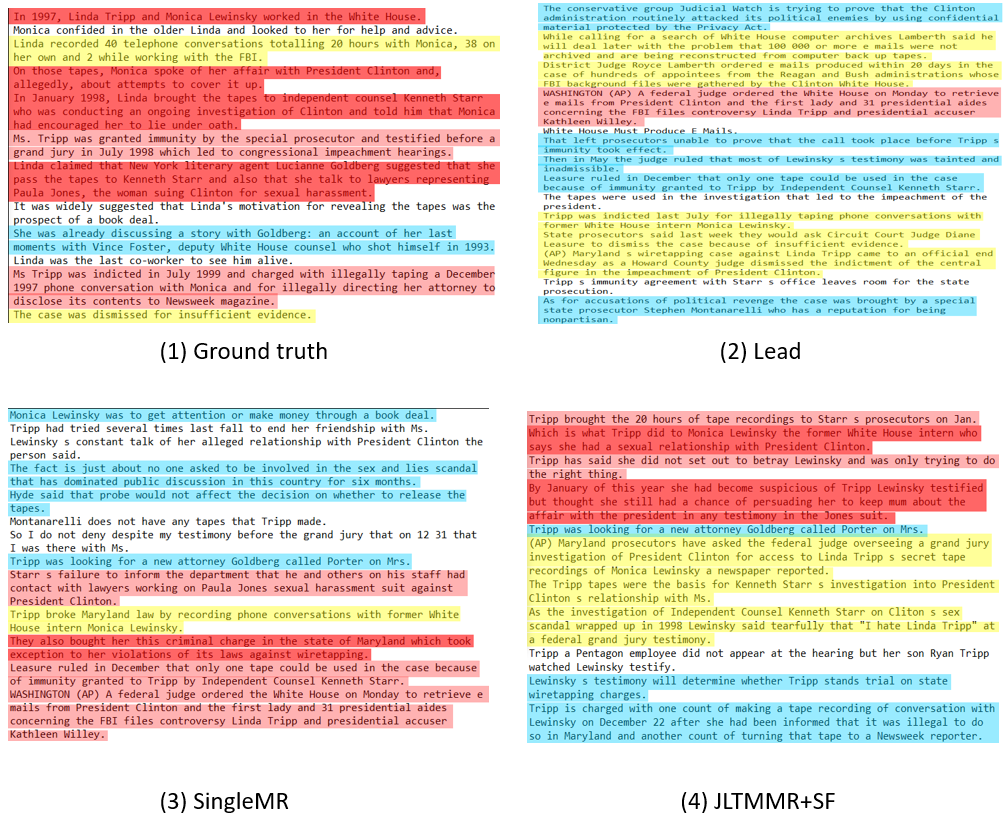}
\caption{Visualization of different models.}
\label{resultshow}
\end{figure}

We present an example to illustrate our methods at work. It is an instance from the DUC 2007 dataset under topic D0731G. From top-left to bottom-right, it is the Ground truth, Lead, SingleMR, JLTMMR+SF in Fig.~\ref{resultshow}. The depth of the color corresponds to the importance of the sentence given by ground truth or models. It is obvious that the JLTMMR+SF model has the best performance.
\subsection{The Effect of Adding Feedback}
The normal processing of machine learning is: (1) dealing with data; (2) training model by using data (with or without tags); (3) predicting unknown data by model. The performance of the model depends on the one shot training because the model will not be changed after training. It will make the model cannot adapt to the development of data. Therefore, some researchers \cite{Stumpf2007Toward} propose that adding feedback to the machine learning model can play an important role in improving the model's performance.

It can improve the performance of document summarization task by adding feedback to the machine learning model. For example, Avinesh et al. \shortcite{Pvs2017Joint} proposed a model for multi-document summarization task by learning from user's feedback. The model can better meet the user's needs by adjusting the weight of ILP (Integer Linear Programming) according to the user's selection of the concept. Xu et al. \shortcite{xu2009user} proposed a model to generate the summary. It can predict the importance of words in document through tracking people's attention when they are reading documents. After that, it can combine the important words together as the abstract of documents. Therefore, we propose a method to simulated adding user feedback to our JLTMMR model.

\begin{figure}[H]
\centering
\includegraphics[width=0.6\textwidth]{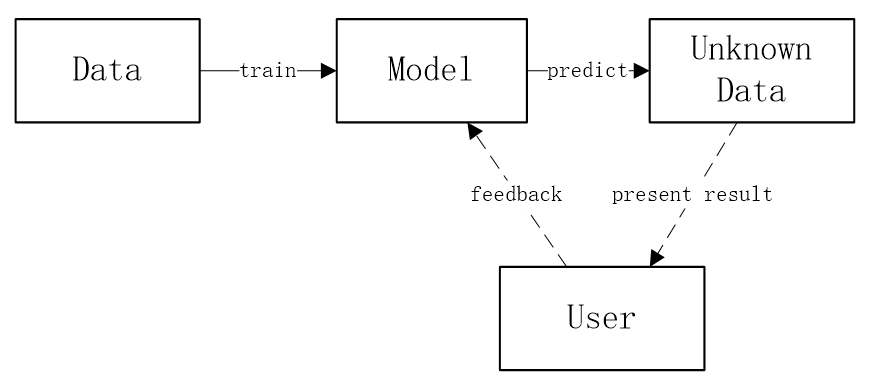}
\caption{Model with feedback}
\label{modelwithfeedback}
\end{figure}

Firstly, user feedback contains three type of behaviors: Like, Normal, Hate. We generate a user feedback dataset including each sentence in DUC 2005 dataset and DUC 2007 dataset, and their ROUGE value. We split the data into three pieces according to their ROUGE value from big to small. There are one piece of data with the largest ROUGE value denotes the Like behavior while the piece of data with the smallest ROUGE value denotes the Hate behavior, and the rest of data represents Normal. After that, we can get 600 feedback of each behavior to our model. Then we use fasttext model \cite{Joulin2016Bag} to train a classification model by using the feedback data. Finally, we can get a classification score by using the classification model on DUC 2006 dataset and combine the manifold ranking score and classification score to generate the final saliency score.

\begin{figure}[H]
\centering
\includegraphics[width=0.8\textwidth]{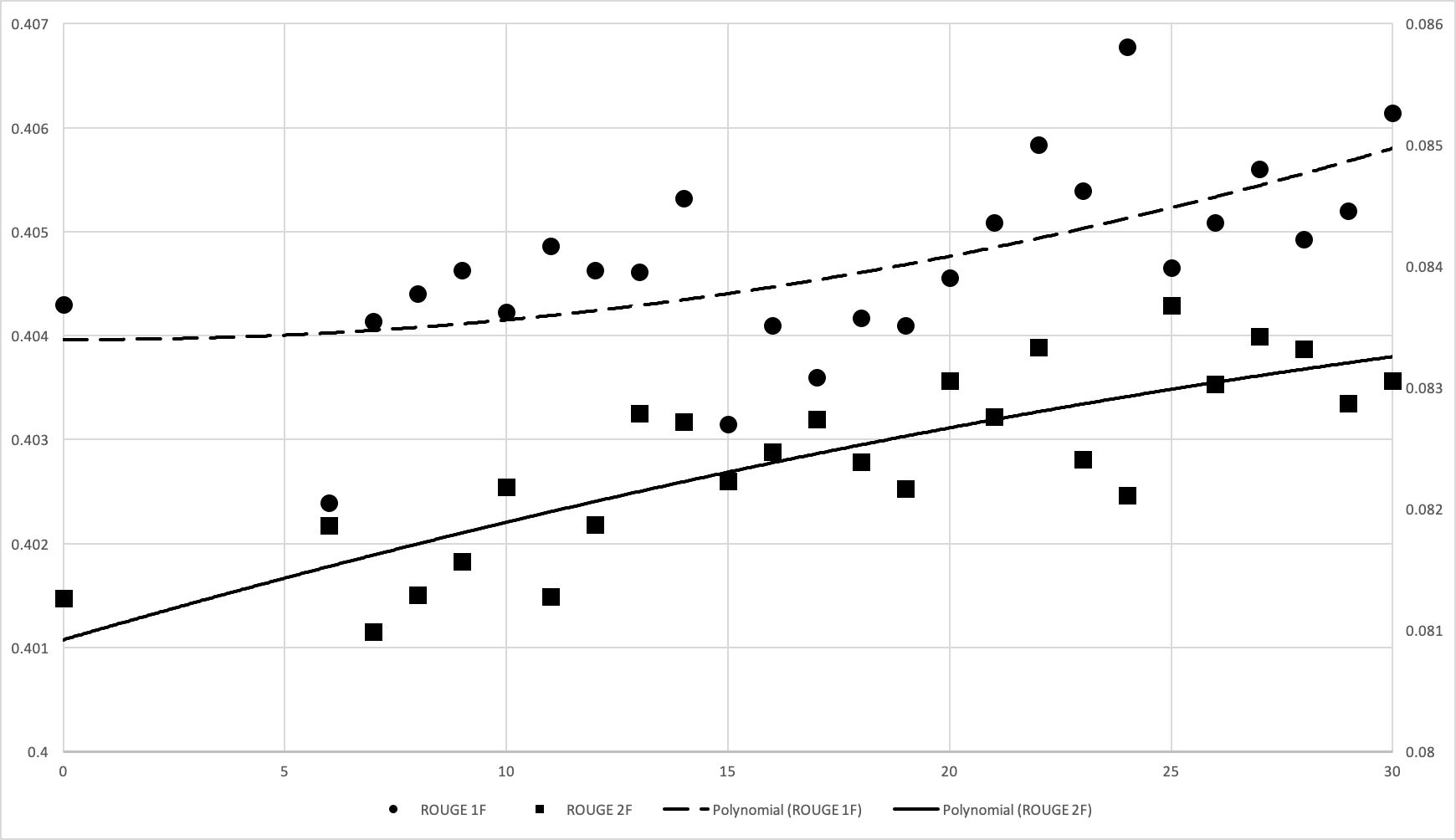}
\caption{The result of feedback}
\label{feedbackresult}
\end{figure}

As shown in Fig.~\ref{feedbackresult}, the effect of user feedback on the summary generation of the DUC 2006 dataset was simulated for a continuous 30-days simulation. Due to the cold start problem, we first accumulated the user feedback data for the previous 5 days, and evaluated the feedback results from the 6th day. It can be seen from the figure that with the continuous accumulation of feedback data by users, the summary effect generated by the final model is also showing an upward trend. Due to the instability of the fasttext model caused by the small amount of data, the summary effect will fluctuate up and down, but according to the polynomial fitting of the ROUGE-1 and ROUGE-2 evaluation indicators, the overall trend is still consistent upward. This can show that the knowledge of using user feedback can play a positive role in lifelong learning.

\section{Conclusions and future directions}
Our main concern in this work was to study the document summarization task. It can help people to read important information in a large amount of documents more comprehensively, quickly and accurately on the Internet. There are many researchers use the manifold ranking method to solve the document summarization task because it has a significant effect on the ranking of unknown data based on known data by using a weighted network. However, their models only consider the original features but ignore the semantic features of sentences when they construct the weighted networks for the manifold ranking method.

We propose a joint model for document summarization task which is called joint lifelong topic model and manifold ranking (JLTMMR) to solve the problem in manifold ranking. Firstly, we propose a lifelong topic model by adding the constraint of knowledge and the constraint of the relationship between documents to the non-negative matrix factorization model. It can improve the quality of the topic and dig out a better document semantic features. Secondly, to solve the document summarization task, we combine the lifelong topic model and manifold ranking method (JLTMMR). After that, we also get another model called JLTMMR+SF by adding statistical features to the JLTMMR model. Experiments show that the JLTMMR model and the JLTMMR+SF model can achieve a better result than other baseline models for multi-document summarization task. At the same time, our models also have a good performance on single document summarization task.

Analyzing all the results in this paper, we clearly understood that adding knowledge to our model can not only improve the quality of the topic but also play a positive effect in the document summarization task. What's more, we find that it can achieve a better performance after adding feedback as knowledge to our model. It shows that the effect of adding knowledge to lifelong machine learning model is significant.

The proposed model has some advantages over the existing document summarization models, but there is still a lot of space to improve for the document summarization task.

(1) There is still a lot of work to do for lifelong machine learning.

(i) Knowledge representation: In this paper, we use a simple way to represent knowledge by using the co-occurrence relationship of words. However, it has shortcoming because there are lots of knowledge representation for human, including knowledge inference, knowledge graph, knowledge transfer and so on. Thus, it can improve the lifelong machine learning by using a better knowledge representation.

(ii) Knowledge extraction: Is the knowledge always right? For example, the knowledge "apple tastes good" is useful for the task related to fruit, but useless and negative for the task related to computer. Thus, it's an improvement for lifelong machine learning by considing how to extract related knowledge from existing knowledge.

(iii) Knowledge transfer: It's a question worth exploring that how to use the current knowledge to carry out knowledge transfer, so that the model can meet the needs of the new task.

(2) For the document summarization task, we always use ROUGE to evaluate the quality of the abstract produced by machine. However, the formula of ROUGE shows that it dosen't care about the deep semantic relations of abstract. The result is completely different when we use different words to express the same meaning. It has a significant effect on the document summarization task so it is also a hot research topic for document summarization.

(3) The extractive method for document summarization is easily to understand, while the abstractive method is more similar to human ways. However, most of the abstractive method are used for sentence compression task. These models are implemented by using CNN + Attention or RNN + Attention, so how to use them to generate a better summary, is also a hot research topic.

\label{lastpage}

\end{document}